\documentclass{article}
\usepackage{amsthm}
\usepackage{amsmath}
\usepackage{amssymb}
\usepackage{epsfig}

\usepackage{graphicx} 
\usepackage{subfigure}

\usepackage{natbib}

\usepackage{algorithm}
\usepackage{algorithmic}

\usepackage{hyperref}

\usepackage[accepted]{icml2013}

\icmltitlerunning{Gradient-based Laplacian Feature Selection}

\begin{document}
\twocolumn[
\icmltitle{Gradient-based Laplacian Feature Selection}

\icmlauthor{Bo Wang}{bo.wang@sickkids.ca}
\icmlauthor{Anna Goldenberg}{anna.goldenberg@sickkids.ca}
\icmladdress{SickKids Research Institute, 101 College St. East Tower, Toronto, ON, M5G 1L7}
\icmlkeywords{boring formatting information, machine learning, ICML}

\vskip 0.3in
]
\begin{abstract}
Analysis of high dimensional noisy  data is of essence across a variety of research fields. Feature selection techniques are designed to find the relevant feature subset that can facilitate classification or pattern detection. Traditional (supervised) feature selection methods utilize label information to guide the identification of relevant feature subsets. In this paper, however, we consider the unsupervised feature selection problem. Without the label information, it is particularly difficult to identify a small set of relevant features due to the noisy nature of real-world data which corrupts the intrinsic structure of the data. Our Gradient-based Laplacian Feature Selection (GLFS) selects important features by minimizing the variance of the Laplacian regularized least squares regression model. With $\ell_1$ relaxation, GLFS can find a sparse subset of features that is relevant to the Laplacian manifolds. Extensive experiments on simulated, three real-world object recognition and two computational biology datasets, have illustrated the power and superior performance of our approach over multiple state-of-the-art unsupervised feature selection methods. Additionally, we show that GLFS selects a sparser set of more relevant features in a supervised setting outperforming the popular elastic net methodology.
\end{abstract}
\section{Introduction}
In many domains including computer vision \cite{Yang:EECS-2007-99}, pattern recognition \cite{duda:hart:stork:2001} and more recently in computational biology \cite{Fan06}, one is faced with analysis of very high dimensional noisy data. The goal of feature selection is to remove the signal-drowning noise and to reduce the dimensionality to make the problem more tractable \cite{miguel03}. In computational biology, feature selection is of particular importance as the signal-to-noise ratio is very small and collecting the data is often very expensive, resulting in a small sample size. New strategies for selecting relevant features are required to cope with these constraints.

Feature selection approaches can be classified into supervised and unsupervised. Supervised methods have been successfully applied in many domains \cite{Guyon:2002,Robnik2003} often incorporating feature selection into classification models where labeled data is used to identify the subset of features that are most predictive of the outcome. A sparse generalized linear model with a penalty function is one of the most popular choices to date~\cite{Tibshirani94, elasticnet}.  In many situations, however, the labels may not be available. For example, in trying to identify biological underpinnings of a disease (a gene-centric signature), it is very important to take into account the disease's heterogeneity, i.e. disease subtypes that may not have been identified yet. Thus, unsupervised methods are needed to enable new discoveries and a deeper understanding of the data.

Unsupervised feature selection methods are usually guided by the structure in the data, e.g. searching for features explaining data partitioning to improve clustering performance~\cite{Constantinopoulos:2006,Dy:2004}. Most of these methods are computationally expensive and do not scale well~\cite{Cai:2010}. Many selection criteria are reviewed in~\cite{liu05}. Popular criteria include Laplacian Score~\cite{LScore05}, designed to evaluate the importance of each feature to the manifold structure of the data and optimizing cluster coherence by spectral gap of the corresponding affinity matrix ~\cite{wolf03}. Sparse PCA (SPCA) is another popular method that attempts to identify a sparse set of features that can capture most of the variance in the data ~\cite{Zou04sparseprincipal}. More recently, a variance minimization criterion based on Laplacian-Regularized Least Squares regression (LapRLS) was proposed~\cite{xiaofei12}. Though based on regression, the parameter covariance matrix does not depend on the label, thus while still unsupervised, this objective function has advantages over previously proposed methods in that it optimizes both discriminative and geometrical properties of the data. However, the method as proposed has several disadvantages. First, the authors used an approximation to the full covariance matrix. Second, since $\ell_0$ optimization is an NP-hard problem, they employed a greedy optimization procedure to approximate the optimal solution. Last but not least, they need to specify the number of desired features $k$ which in most problems is not known a priori. Thus, while we believe that \cite{xiaofei12} was on the right track, it stopped short of delivering a scalable method suitable for a wide use in practice.

In this paper we propose an unsupervised feature selection method called Gradient-based Laplacian-regularized Feature Selection (GLFS) by defining a variance minimizing objective function and an optimization strategy to solve it. Similarly to \cite{xiaofei12},  GLFS is based on LapRLS \cite{LapRLS06}, however GLFS has several significant advantages over \cite{xiaofei12}: 1) we use exact expression of the full covariance matrix which properly exploits the inner manifold structure of the data; 2) instead of greedily solving the $\ell_0$ norm problem, we propose a relaxation to $\ell_1$ which offers a desirably sparse solution and greatly improves computational power; 3) we provide a closed-form first-order Jacobian matrix to make the $\ell_1$ optimization amenable to most Newton-type sparse solvers; 4) a line search technique is proposed to fit the penalty parameter automatically. \comment{We show that our GLFS results in \emph{significant} rather than incremental improvements  over  \cite{xiaofei12}.} We perform extensive experiments comparing GLFS to  \cite{xiaofei12}, Laplacian Score \cite{LScore05}, the method of \citep{wolf03} and SPCA ~\cite{Zou04sparseprincipal} on the simulated data with very few informative and thousands of irrelevant features. We further show that our performance is superior on MNIST,COIL20 and AT$\&$T object recognition datasets. We show a significant increase in unsupervised subtype detection in glioblastoma (GBM) from the expression data obtained from The Cancer Genome Atlas (TCGA). Finally, we show that even in the supervised setting extended GLFS outperforms some of the most commonly used classifiers with feature selection (such as Elastic net~\cite{elasticnet}) on the Leukaemia microarray dataset. Our results show that by using the exact formulation of the variance in LapRLS regression setting and the $\ell_1$ relaxation of the constraint we are able to provide a much more robust and scalable unsupervised feature selection method than those are currently available in the literature.
				
\section{Method}
\subsection{Problem Setting}
Let $\mathbf{X} = (\mathbf{x}_1,\ldots, \mathbf{x}_n)$ be a $d\times n$ input data matrix, where $d$ is the number of features and $n$ is the number of data samples. The $i$-th row of $\mathbf{X}$ is denoted as $\mathbf{f}_i^T\in \mathbb{R}^n$, each corresponding to a feature. The goal of feature selection is to choose the most informative feature subset $\mathcal{G}=\{\mathbf{f}_{i_1},\mathbf{f}_{i_2},\ldots, \mathbf{f}_{i_k}\}$. Let $\mathbf{X}^{\mathcal{G}}$ be a new data matrix containing only features in $\mathcal{G}$ and $y_i$ be a label of the data point $i$ (the labels are not really given but we treat them as such for the ease of derivation). We consider a linear regression model in the space of selected features $\mathcal{G}$:
\begin{equation}
y = \mathbf{w}^T\mathbf{x}^{\mathcal{G}} + \epsilon,
\end{equation}
where $\epsilon$ is a random noise with zero mean and variance $\sigma^2$ and $y$ is the target.
By incorporating the Laplacian regularizer into the sum squared error, the objective function becomes:
\begin{align}
\label{eqn:objLapRLS}
J(\mathbf{w}) =\sum_{i=1}^n(\mathbf{w}^T\mathbf{x}_i^{\mathcal{G}}-y_i)^2 &+ \frac{\lambda_1}{2}\sum_{i,j = 1}^n(\mathbf{w}^T\mathbf{x}_i^{\mathcal{G}}-\\ \notag&-\mathbf{w}^T\mathbf{x}_j^{\mathcal{G}})^2S_{ij} + \lambda_2\|\mathbf{w}\|^2_2,
\end{align}
where $\mathbf{S}$ is the similarity matrix whose entry $S_{ij}$ represents the similarity/affinity between $\mathbf{x}_i$ and $\mathbf{x}_j$.
The Laplacian regularized regression model effectively takes the data manifold into consideration and avoids overfitting ~\cite{LapRLS06}.

If we define the Laplacian matrix as $L = D - S$, where $D$ is a diagonal matrix with $D_{ii} = \sum_{j=1}^nS_{ij}$, then there is a closed-form solution of Eq \ref{eqn:objLapRLS} as is indicated in ~\cite{LapRLS06}:
\begin{equation}
\label{eqn:expW}
\mathbf{w} = (\mathbf{X}^{\mathcal{G}}(\mathbf{X}^{\mathcal{G}})^T + \lambda_1\mathbf{X}^{\mathcal{G}}L(\mathbf{X}^{\mathcal{G}})^T + \lambda_2I)^{-1}\mathbf{X}^{\mathcal{G}}\mathbf{y},
\end{equation}
where $I$ is an identity matrix. If we define $Z = \mathbf{X}^{\mathcal{G}}(\mathbf{X}^{\mathcal{G}})^T + \lambda_1\mathbf{X}^{\mathcal{G}}L(\mathbf{X}^{\mathcal{G}})^T + \lambda_2I$, we have $\mathbf{w}= Z^{-1}\mathbf{X}^{\mathcal{G}}\mathbf{y}$. Since $Cov(\mathbf{y}) = \sigma^2I$ and $Z$ is symmetric, the covariance matrix of $\mathbf{w}$ is
\begin{eqnarray}
Cov(\mathbf{w}) &=& Z^{-1}\mathbf{X}^{\mathcal{G}}Cov(y)(\mathbf{X}^{\mathcal{G}}
)^TZ^{-1}\nonumber\\
&=& \sigma^2Z^{-1}\mathbf{X}^{\mathcal{G}}(\mathbf{X}^{\mathcal{G}})^TZ^{-1}.
\end{eqnarray}

\subsection{Feature Selection via Variance Minimization }

Let's introduce $d$ indicator variables $\beta_i$ to be $1$ if the $i$-th feature $\mathbf{f}_i\in \mathcal{G}$, $0$ otherwise. Let $\mathbf{\beta}^T = (\beta_1,\ldots,\beta_n)$. Then we have $\|\mathbf{\beta}\|_0 = |\mathcal{G}|$ and
$$(\mathbf{X}^{\mathcal{G}})^T\mathbf{X}^{\mathcal{G}} = \sum_{i=1}^{d}\beta_i\mathbf{f}_i\mathbf{f}_i^T =  \mathbf{X}^T\mathbf{diag}(\mathbf{\beta})\mathbf{X}$$

Our feature selection criterion selects significant features that minimize the covariance of $\mathbf{w}$. The motivation behind this criteria is two-fold: 1) the covariance of $\mathbf{w}$ measures the robustness of the regression model, minimizing the covariance then facilitates the selection of features that contribute most to the manifolds of the whole data; 2) the covariance of $\mathbf{w}$ reflects the fitness of the regression model on the task of classification. This property in turn means our objective of minimizing the covariance can choose those features that are signatures of clusters in the data with potential to benefit the further classification task.

If a desired number of selected features is $k$, then our objective function formulation would be
\begin{eqnarray}
\label{eqn:objFS}
\min Tr(Cov(\mathbf{w})) &=& Tr(\sigma^2Z^{-1}\mathbf{X}\mathbf{diag}(\mathbf{\beta})\mathbf{X}^TZ^{-1})\nonumber\\
s.t. &&\quad\quad\|\beta\|_0 = k, \beta_i \in \{0,1\}.
\end{eqnarray}

Note that, by minimizing the covariance of $\mathbf{w}$, we do not use any information of the label $\mathbf{y}$, which is only a hypothetical label and plays no role in the feature selection criterion. Therefore, even though we began with a supervised method --regression, our feature selection method is completely unsupervised.

The matrix $Z$ can be rewritten as follows by using the Woodbury formula:
$$
Z^{-1} = \frac{1}{\lambda_2}I-\frac{1}{\lambda_2^2}\mathbf{X}^{\mathcal{G}}((I+\lambda_1L)^{-1} + \frac{1}{\lambda_2}(\mathbf{X}^{\mathcal{G}})^T\mathbf{X}^{\mathcal{G}})^{-1}(\mathbf{X}^{\mathcal{G}})^T.
$$
Let $M = \lambda_2(I + \lambda_1L)^{-1}$, then we have
$$
Z^{-1} = \frac{1}{\lambda_2}I-\frac{1}{\lambda_2}\mathbf{X}^{\mathcal{G}}(M + (\mathbf{X}^{\mathcal{G}})^T\mathbf{X}^{\mathcal{G}})^{-1}(\mathbf{X}^{\mathcal{G}})^T.
$$

Hence the objective function in (\ref{eqn:objFS}) becomes:
\begin{eqnarray}
Tr(Cov(\mathbf{w})) &=& \sigma^2Tr(Z^{-1}\mathbf{X}^{\mathcal{G}}(\mathbf{X}^{\mathcal{G}})^TZ^{-1})\nonumber\\
&=& \frac{\sigma^2}{\lambda_2^2}Tr((\mathbf{X}^{\mathcal{G}})^T[I-\mathbf{X}^{\mathcal{G}}(M\nonumber \\
&+& (\mathbf{X}^{\mathcal{G}})^T\mathbf{X}^{\mathcal{G}})^{-1}(\mathbf{X}^{\mathcal{G}})^T]^2\mathbf{X}^{\mathcal{G}})\nonumber\\
&=& \frac{\sigma^2}{\lambda_2^2}Tr(A-2A(M+A)^{-1}A\nonumber\\
&+&A(M+A)^{-1}A(M+A)^{-1}A)\nonumber\\
&=& \frac{\sigma^2}{\lambda_2^2}Tr(A[(M+A)^{-1}M]^2)\nonumber\\
&=& \frac{\sigma^2}{\lambda_2^2}Tr(A(M+A)^{-2}M^2)\nonumber\\
\end{eqnarray}
where $A = (\mathbf{X}^{\mathcal{G}})^T\mathbf{X}^{\mathcal{G}} = \mathbf{X}^T\mathbf{diag}(\mathbf{\beta})\mathbf{X}$.
Note that here we repeatedly use the fact that both $Z$ and $Z^{-1}$ are symmetric, and some trace properties: $Tr(AB) = Tr(BA)$ and $Tr(ABC) = Tr(ACB)$ if $A$,$B$ and $C$ are symmetric. Hence, the optimization problem in eqn(\ref{eqn:objFS}) can be reformulated as
\begin{eqnarray}
\label{eqn:objFS1}
minimize_{\beta} \mathcal{Q}&=&Tr(A(M+A)^{-2}M^2)\nonumber\\
s.t. &&\quad\quad A = \mathbf{X}^T\mathbf{diag}(\mathbf{\beta})\mathbf{X}\nonumber\\
     &&\|\beta\|_0 = k, \beta_i \in \{0,1\}.
\end{eqnarray}

Note that we are using an exact formulation of the covariance while ~\cite{xiaofei12} approximates the term $Z^{-1}\mathbf{X}\mathbf{diag}(\mathbf{\beta})\mathbf{X}^TZ^{-1}$ in (\ref{eqn:objFS}) with $Z^{-1}$. This approximation disregards an important contribution of the variance in the data with respect to the sparse set of informative features leading to inferior performance as is illustrated in our Experiments section.

\subsection{Optimization}

Since it is NP-hard to solve the $\ell_0$ constrained problem directly in addition to not knowing the value of $k$ in advance, we propose to use a sparse $\ell_1$ constraint transforming the problem in (\ref{eqn:objFS1}) into
\begin{eqnarray}
\label{eqn:objFS2}
minimize_{\mathbf{\beta}} \mathcal{J}&=&Tr(A(M+A)^{-2}M^2)+ \lambda\|\mathbf{\beta}\|_1\nonumber\\
s.t. && A = \mathbf{X}^T\mathbf{diag}(\mathbf{\beta})\mathbf{X}\nonumber\\
     &&\mathbf{\beta}\succcurlyeq 0 .
\end{eqnarray}
where $\lambda$ is a regularization parameter that can be determined either from prior information or validation data.
This is a typical $\ell_1$ non-convex optimization problem.
Note that we relaxed the constraint that $\mathbf{\beta}$ must be either $0$ or $1$. This allows us to interpret $\mathbf{\beta}$ as a feature weight vector to prioritize features and also facilitates the optimization solving. The idea behind this relaxation is due to two reasons: 1) $\ell_1$ relaxation provides a sparse solution and removes the necessity to either manually or a priori set the number of desired features $k$ which is usually unknown; 2) compared with some greedy algorithms for this optimization ~\cite{xiaofei12} to solve the $\ell_0$ constraint problem directly, our $\ell_1$ relaxation is more efficient and faster to solve (see Complexity Analysis section).
\subsubsection{First-Order Jacobian}

From (\ref{eqn:objFS1}) $\mathcal{Q} = A(M+A)^{-2}M^2$, where $A = \sum_{i=1}^{d}\beta_i\mathbf{f}_i\mathbf{f}_i^T =\mathbf{X}^T\mathbf{diag}(\mathbf{\beta})\mathbf{X}$. The first-order derivative $\frac{\partial\mathcal{Q}}{\partial\beta_i}$ is then given by:
\begin{eqnarray}
\frac{\partial\mathcal{Q}}{\partial\beta_i}&=&\frac{\partial tr(A(M+A)^{-2}M^2)}{\partial\beta_i}\nonumber\\
&=&tr(\frac{\partial (A(M+A)^{-2}M^2)}{\partial\beta_i})\nonumber\\
&=&tr((\frac{\partial A}{\partial\beta_i})^T\frac{\partial (A(M+A)^{-2}M^2)}{\partial A})\nonumber\\
&=&\mathbf{f}_i^T(\nabla_{A}tr(A(M+A)^{-2}M^2))\mathbf{f}_i\nonumber\\
&=&\mathbf{f}_i^T[(M+A)^{-2}-2(M+A)^{-3}A]^TM^2\mathbf{f}_i\nonumber\\
&=&\mathbf{f}_i^T(M-A)(M+A)^{-3}M^2\mathbf{f}_i.
\end{eqnarray}
In the derivation, we repeatedly use the symmetric and positive semi-definite properties of $M$ and $A$. Also, $\frac{d}{dt}tr g(X+tY) = tr \frac{d}{dt}g(X+tY)$, $\nabla_Xg(f(X)^T) = \nabla_Xf^T\nabla_fg$, $\nabla_Xg(f(X)^T,h(X)^T) = \nabla_Xf^T\nabla_fg + \nabla_Xh^T\nabla_hg$ (see ~\cite{Dattorro2011}).

The calculation of the first-order Jacobian allows the use of many Newton-type optimization solvers for $\ell_1$ sparse system. We use the Orthant-Wise descent~\cite{Andrew:2007} which is a recently proposed sparse solver that consists of iterations taking the form
$$
\mathbf{\beta} \leftarrow \mathcal{P}_{\mathcal{O}}[\beta-\lambda\mathcal{P}_{\mathcal{S}}[H^{-1}\nabla_{\mathcal{\beta}}\mathcal{Q}]].
$$
The two projection operators $\mathcal{P}_{\mathcal{S}}$ and $\mathcal{P}_{\mathcal{O}}$ are the key advantages of the Orthant-Wise descent. The operator $\mathcal{P}_{\mathcal{S}}$ sets elements of the vector $H^{-1}\nabla_{\mathcal{\beta}}\mathcal{Q}$ to zero if they do not have the same sign as the corresponding element of $\nabla_{\mathcal{\beta}}\mathcal{Q}$. By that, it guarantees the Netwon-like direction to be a descent direction. Similarly to the active set method, this is equivalent to setting the working set to those variables whose sign does not change after applying the inverse-Hessian scaling. The other operator, $\mathcal{P}_{\mathcal{O}}$, sparsifies the current solution by projecting the step onto the orthant containing $\mathbf{\beta}$. These properties are helpful in our framework because the Orthant-Wise descent can make a large number of changes to those non-zero variables and these projectors can help distinguish the key features that have large effect on the variance matrix defined in Eqn \ref{eqn:objFS2}.

\subsubsection{Parameter Search}
We note that while our relaxation leads to a vector of zeros be the optimal solution of Eqn \ref{eqn:objFS2}, this is a common and successful technique in numerical optimization ~\cite{nocedal99}, where the desirable solution is a `good' local minimum. This property makes the problem sensitive to the value of $\lambda$. If it is too big, the $\ell_1$ solver will quickly degenerate to a vector of zeros; while if it is too small, the solution will lead to a dense local optimum. This is a common problem in many applications (e.g., active contour problem ~\cite{activecontour}). We thus propose a line search technique to optimize $\lambda$.  First, we select a small value $\lambda_0$ and a large positive constant $C$. At iteration $t$, if $\mathcal{Q}(\lambda_t)=0$, i.e., $\mathbf{\beta} = \mathbf{0}$, we update $\lambda_{t+1} = \frac{1}{2}\lambda_t$, otherwise $\lambda_{t+1} = \frac{C}{2^t}\lambda_t$. So our maximum iteration number would be $t_{max} = \lfloor\log_2C\rfloor$. Also, we adopt the strategy of ``early stopping''~\cite{Erhan2010} by keeping track of the value $Q(\lambda_t)$. If we encounter the case $Q(\lambda_{t-1})=Q(\lambda_{t+1})=0$, we stop and use the value of $\lambda_t$. Early stopping helps us to reach a sparse local optimum quickly and efficiently.

\subsubsection{Complexity Analysis}
The proposed method consists of two main steps: calculation of the derivatives and an update of the desired variables. If we assume the iteration number is $T$, then it is easy to verify that the time needed to obtain a desired value of $\mathbf{\beta}$ is $\mathcal{O}(Tn^2d)$, i.e. it is linear in the number of features. The convex optimization in ~\cite{xiaofei12} needs $\mathcal{O}(n+d^2)^2d^{2.5}$ calculations, whereas our $\mathcal{O}(Tn^2d)$ is just the worst-case complexity and our real calculation cost is much less. To be specific, if we denote the average number of non-zero values in $\mathbf{\beta}$ as $d^{+}$, the average time complexity is $\mathcal{O}(Tn^2d^{+})$.

\section{Experiments}
In this section, we test GLFS on a variety of tasks. We compare the proposed method to several existing unsupervised feature selection methods: 1)The LapAOFS algorithm~\cite{xiaofei12}; 2) The LapDOFS algorithm~\cite{xiaofei12}; 3) Laplacian Score(LS) ~\cite{LScore05}; 4) $Q-\alpha$ algorithm~\cite{wolf03}; 5) Sparse Principle Component Analysis (SPCA)~\cite{Zou04sparseprincipal}.
The LapAOFS and LapDOFS aim to find the best $k$ features that minimize approximate variance of the Laplacian regularized regression. The two methods differ in the objective function: LapAOFS tries to minimize the trace of the covariance matrix while LapDOFS tries to minimize the determinant of the covariance matrix. The Laplacian Score tries to sort all the features by the importance to the Laplacian structure. The $Q-\alpha$ algorithm aims to maximize the cluster coherence. The SPCA searches for a sparse set of features, whose linear combination can explain most of the variation in the data ~\cite{elasticnet}. Note that, for LapAOFS and LapDOFS, we use the greedy optimization technique rather than semi-definite convex optimization as is described in the original paper \cite{xiaofei12}.
In respect to the construction of the Laplacian graph $L$, we use the same method as in ~\cite{LScore05}.
\subsection{Simulations}
In this section, we test the robustness of GLFS to sample noise and the number of irrelevant features on simulated data. We generate a dataset consisting of $400$ samples based on $4$ significant features ($\mathbf{f}_i$, $i=1,2,3,4$). We also add $N$ purely Gaussian noise features with standard deviation $\sigma$. We compare the power of the proposed method to competitors based on a simple score which indicates whether the true $4$ features were ranked among the first $4$:
$$
Score = \frac{1}{4}\sum_{i=1}^4\frac{1}{\max\{4,\mathcal{I}(\mathbf{f_i})\}-3}
$$
where $\mathcal{I}(\mathbf{f}_i)$ is the index of the $i$-th feature when we sort the weights in the decreasing order. It is easy to verify that when the four significant features($\mathbf{f}_i$, $i=1,2,3,4$) have the highest weights, the $Score$ is 1.

\begin{figure}[!htp]
\begin{center}
\begin{tabular}{cc}
\includegraphics[width=0.50\linewidth]{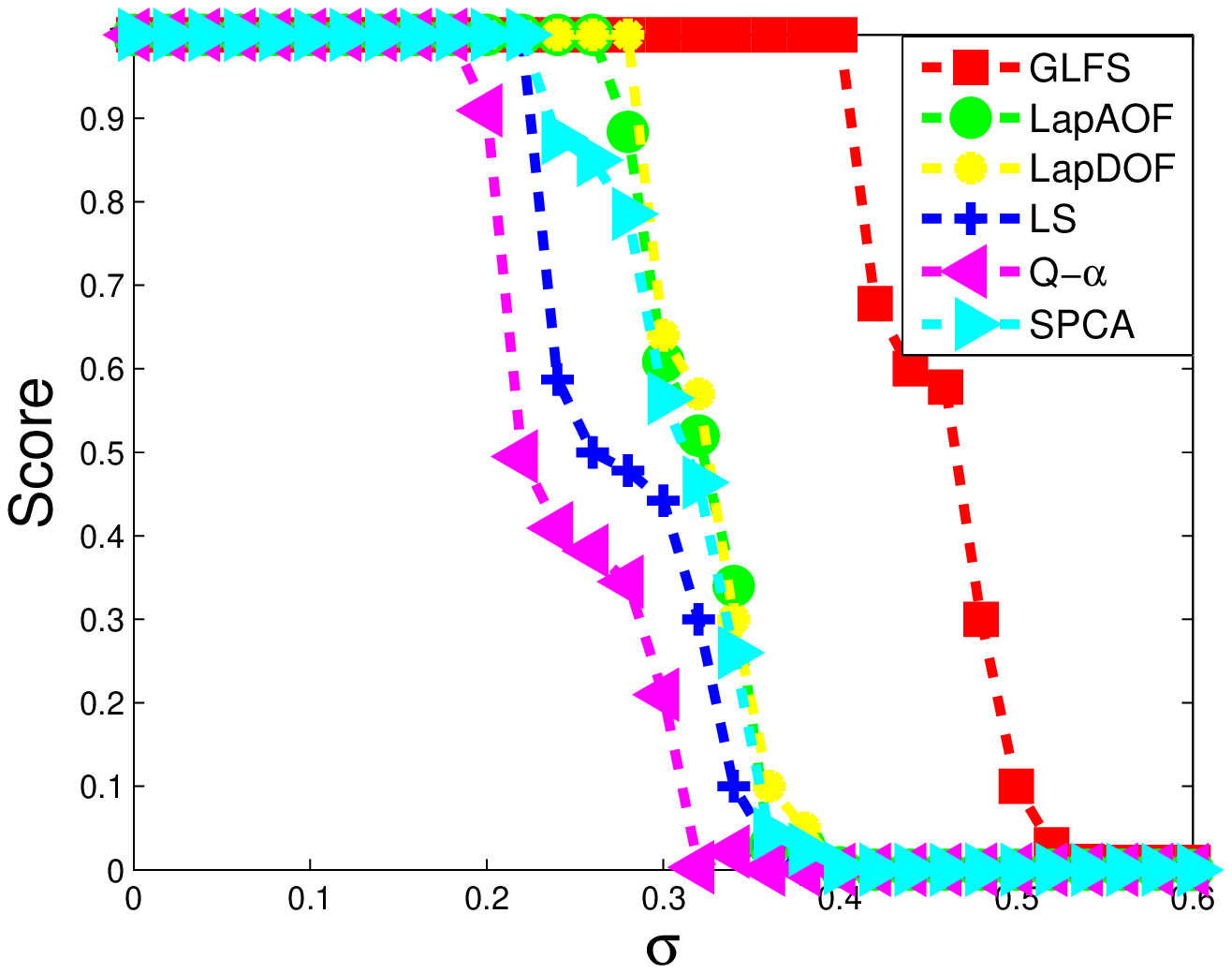} &
\hspace{-8mm} \includegraphics[width=0.50\linewidth]{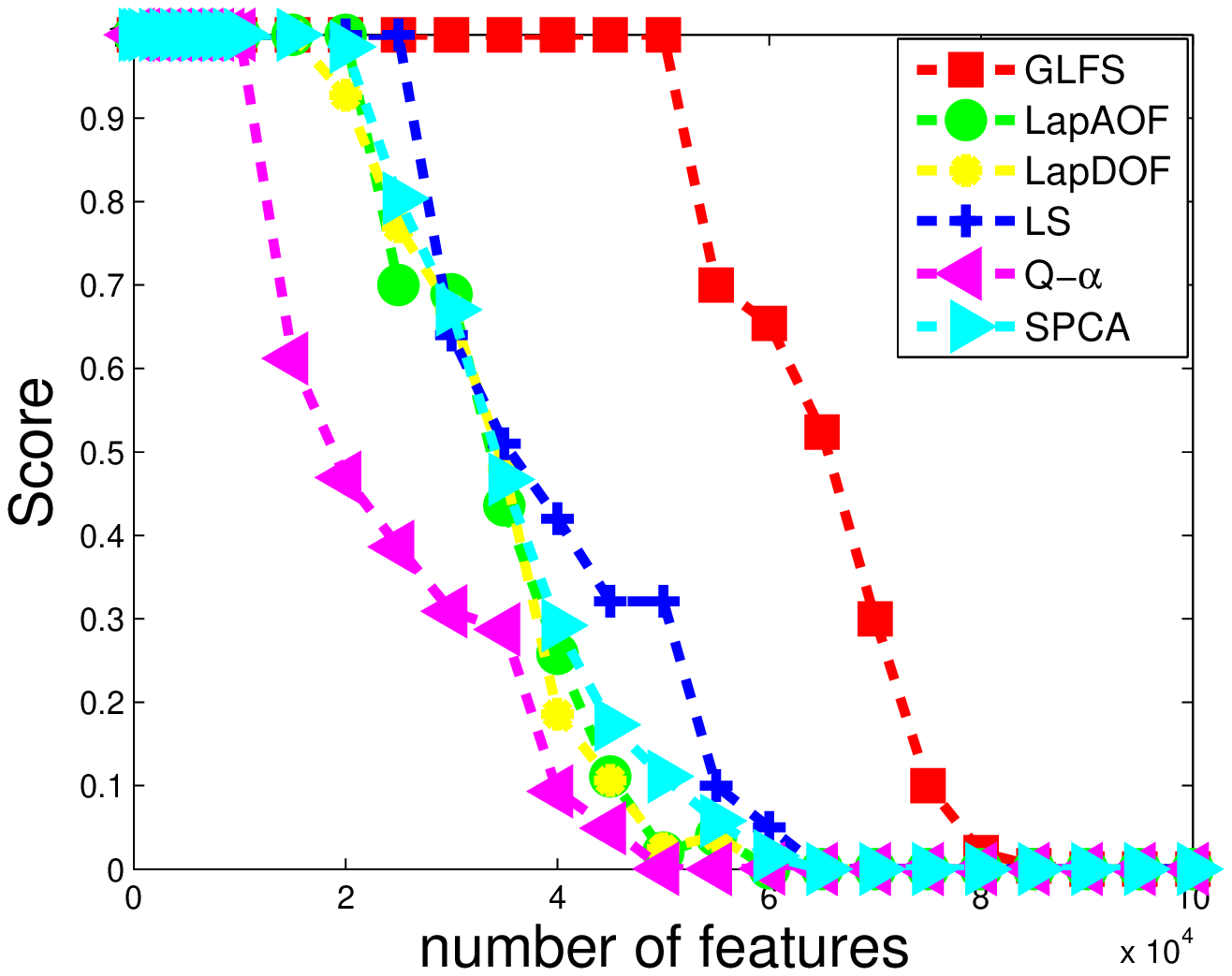} \\
(A)&(B)\\
\end{tabular}
\end{center}
\vspace{-3mm}
\caption{Simulation results. Both plots show the accuracy of recovering the 4 true features as a function of (A)  scale of Gaussian noise when the number of irrelevant features is 1000; (B) the number of irrelevant features when $\sigma = 0.2$.}
\vspace{-3mm}
\label{fig:simunoise}
\end{figure}

First, we fix the number of features ($1000$ irrelevant features with $4$ relevant ones), while varying the standard deviation of the Gaussian noise $\sigma$. We plot the score versus a range of $\sigma$ in Fig.(\ref{fig:simunoise})(A). We can see that GLFS is significantly more robust to noise than other methods, identifying the four true features as the most significant ones with up to $\sigma = 0.415$. Second, we fix the scale of noise to $\sigma=0.2$, while changing the number of irrelevant features and calculating the $Score$ (See Fig.(\ref{fig:simunoise})(B)), obtaining $100\%$ accuracy with up to 60,000 irrelevant features (i.e. 4:60,000 signal-to-noise ratio), significantly higher than any other method. The next best method -- Laplacian Score -- recovers the true set of features in the presence of 30,000 -- half of the irrelevant features.  Interestingly, while LapAOFS and LapDOFS are more robust than LS, SPCA and $Q-\alpha$ to the levels of Gaussian noise in the data, the Laplacian Score is more robust than than the rest in terms of the presence of irrelevant features. This maybe due to the greedy optimization method used in LapAOFS and LapDOFs, making them sensitive to the noise in the eigen structure of the data.

\subsection{Object Recognition}
We applied GLFS to three object recognition benchmark datasets: 1) MNIST handwritten digit database\footnote{http://yann.lecun.com/exdb/mnist/}, which has a training set of 60,000 images and a testing set of 10,000 images. We extracted the first 1,000 images from the training set and the first 1,000 images from the test set. Each image is $32\times 32$ pixels, i.e. each image is represented by a 1024-dimensional vector; 2) the COIL20 image libary\footnote{http://www.cs.columbia.edu/CAVE/software/softlib/ coil-20.php} from Columbia contains 20 classes of objects each consisting of 72 images. The size of each image is $32\times 32$ pixels, hence we have $1440$ $1024$-dimensional vectors; 3) the AT\&T face database\footnote{http://www.cl.cam.ac.uk/research/dtg/attarchive/ facedatabase.html} has a total of 40 subjects (10 samples per subject) with different variations of facial expressions (e.g. glasses or no glasses). Similarly, the size of each image is $32\times32$.

First, we test the effectiveness of the unsupervised feature selection methods by performing k-means clustering in the selected feature space and comparing the clusters to the known class labels in each dataset. We randomly chose $k$ clusters (as indicated in Table \ref{tab:nmiresult}) and used the corresponding data sample to select a $100$ features with each of the competing methods. $k$-means was applied with 10 different random starting points. We use Normalized Mutual Information(NMI)~\cite{Xu03} to measure the performance of the clustering. We ran this experiment 20 times resulting in a total of 200 independent runs. The average performance scores (NMI) are reported in Table \ref{tab:nmiresult}, with the statistically significantly better algorithms indicated in bold (statistical significance was established using t-test with $\alpha=0.05$).

\begin{table*}[!ht]
\centering
\caption{The k-means clustering performance in terms of average (over 200 independent samples) NMI score in the space of 100 best features on three benchmark object recognition datasets. Statistically significantly better result(s) are indicated in bold (according to the t-test $\alpha=0.05$).}
\begin{tabular}{|c|c|c|c|c|c|c|c|c|c|c|}
\hline
&\multicolumn{4}{|c|}{MNIST database}&\multicolumn{3}{|c|}{COIL20 database}&\multicolumn{3}{|c|}{AT$\&$T database}\\
&\scriptsize{$k=3$}&\scriptsize{$k=5$}&\scriptsize{$k=7$}&\scriptsize{$k=9$}&\scriptsize{$k=5$}&\scriptsize{$k=10$}&\scriptsize{$k=15$}&\scriptsize{$k=10$}&\scriptsize{$k=20$}&\scriptsize{$k=30$}\\\hline
\bf{GLFS}&{\bf .596}&{\bf.506}&{\bf .483}&{\bf.482}&{\bf.792}&{\bf.789}&{\bf.778}&{\bf.756}&{\bf.740}&{\bf.750}\\
LapDOFS&.590&.493&{\bf .483}&.479&{\bf .789}&.777&.770&.739&{\bf .737}&.735\\
LapAOFS&.569&.504&{\bf .484}&.474&.774&.785&.776&.728&.735&.736\\
LS&.471&.399&.387&.393&.735&.717&.695&.728&.690&.684\\
$Q-\alpha$&.358&.360&.372&.376&.674&.684&.669&.629&.649&.661\\
$SPCA$&.504&.487&.379&.468&.764&.707&.689&.665&.703&.721\\\hline
\end{tabular}
\label{tab:nmiresult}
\end{table*}

From Table \ref{tab:nmiresult} we can see that variance minimization approaches perform uniformly statistically significantly better than other unsupervised methods tested. SPCA mostly outperforms LS and $Q-\alpha$ but is not on par with GLFS. Furthermore, out of variance minimization methods, GLFS performs as well or in most cases significantly better than LapDOFS and LapAOFS. LapAOFS outperforms LapDOFS on MNIST for $k<9$ but performs worse than LapDOFS on AT\&T dataset, showing no clear winner between trace and determinant objective function optimization.

To further test the effectiveness of the chosen features, we perform ``Leave-One-Out'' classification test. To compare to LapAOFS and LapDOFS fairly, we consider the nearest neighbor classifier, the same classification setting used in ~\cite{xiaofei12}.
The better feature selection method is expected to find features more relevant to classification resulting in higher classification accuracy. For each data point $x_i$, we find its nearest neighbor $x\prime_i$. Let $c(x_i)$ be the class label of $x_i$. Then the classification accuracy is defined as
\begin{equation}
Accuracy = \frac{1}{n}\sum_{i=1}^n\delta(c(x_i),c(x\prime_{i})).
\label{eq:accuracy}
\end{equation}
where $\delta(a,b) = 1$ if $a=b$ and $0$ otherwise. Similarly, we chose $100$ best (or highest weighted) features in each method to test the performance. We compare method performances in Table \ref{tab:nnresult}.

\begin{table}[!ht]
\centering
\caption{The nearest neighbor classification accuracy using $100$ selected features. Statistically significantly better methods according to the t-test ($\alpha =0.05$) are indicated in bold.}
\begin{tabular}{|c|c|c|c|}
\hline
Methods&MNIST&COIL20&AT$\&$T\\
& database& database& database\\
\hline
\bf{GLFS}&{\bf 0.896}&{\bf 1.000}&{\bf 0.940}\\
LapDOFS&{\bf 0.896}&{\bf 1.000}&0.903\\
LapAOFS&{\bf 0.893}&{\bf 1.000}&0.893\\
Laplacian Score&0.749&0.877&0.863\\
$Q-\alpha$&0.850&0.991&0.840\\
$SPCA$&0.821&0.924&0.876\\\hline
\end{tabular}
\label{tab:nnresult}
\end{table}

On the leave-one-out task for MNIST and COIL datasets Laplacian regularized variance minimization methods again statistically significantly outperform competitors. We do show significant improvement over all alternatives in classification accuracy on the AT\&T dataset, where robustness to noise is an important advantage.

\vspace{5mm}
One of the important application areas of feature selection, particularly selecting a small relevant feature set out of a large number of noisy and irrelevant features, is computational biology. Here we show two examples of applying GLFS in the field of CompBio. First, a purely unsupervised feature selection for glioblastoma (GBM) mRNA expression data, using the glioblastoma sub-types as true class labels \cite{Verhaak:2010}. We further test GLFS against an extremely popular elastic net \cite{elasticnet} - a supervised feature selection task on the leukaemia dataset (provided in the original elastic net publication).

\subsection{Glioblastoma Subtypes Identification}

We used the glioblastoma data available from the TCGA repository \cite{TCGA} including expression data for 18,632 genes and $173$ patients processed on array HG-U133A contributed by the Broad Institute. We have then downloaded the information about 4 glioblastoma subtypes (Proneural, Neural, Classical, and Mesenchymal) for the same $173$ patients from the original glioblastoma sub-type article data repository \cite{Verhaak:2010}. We performed no additional pre-processing on the data.

\begin{figure}[!htp]
\vspace{-3mm}
\begin{center}
\begin{tabular}{cc}
\includegraphics[width=0.5\linewidth]{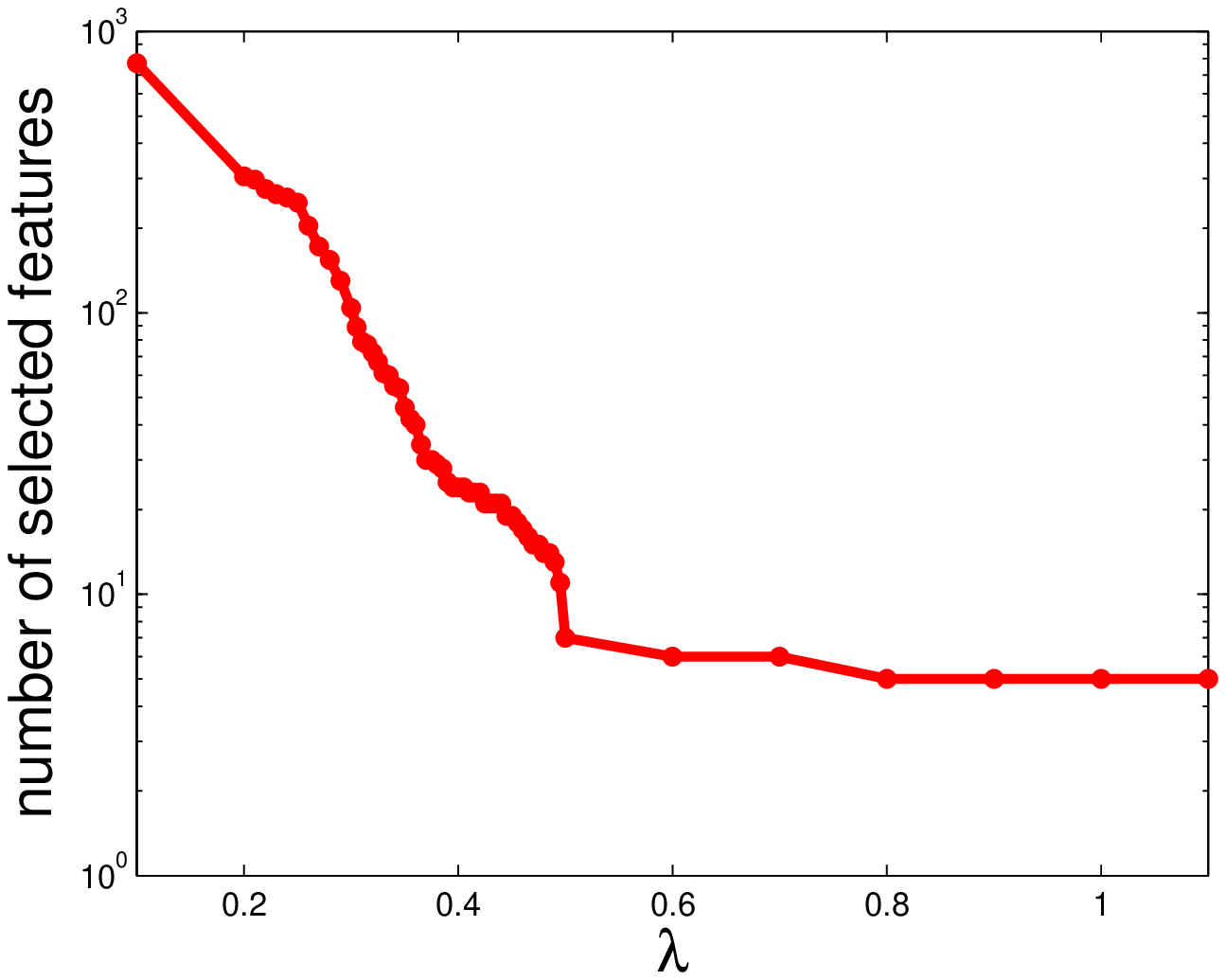} &
\hspace{-8mm} \includegraphics[width=0.5\linewidth]{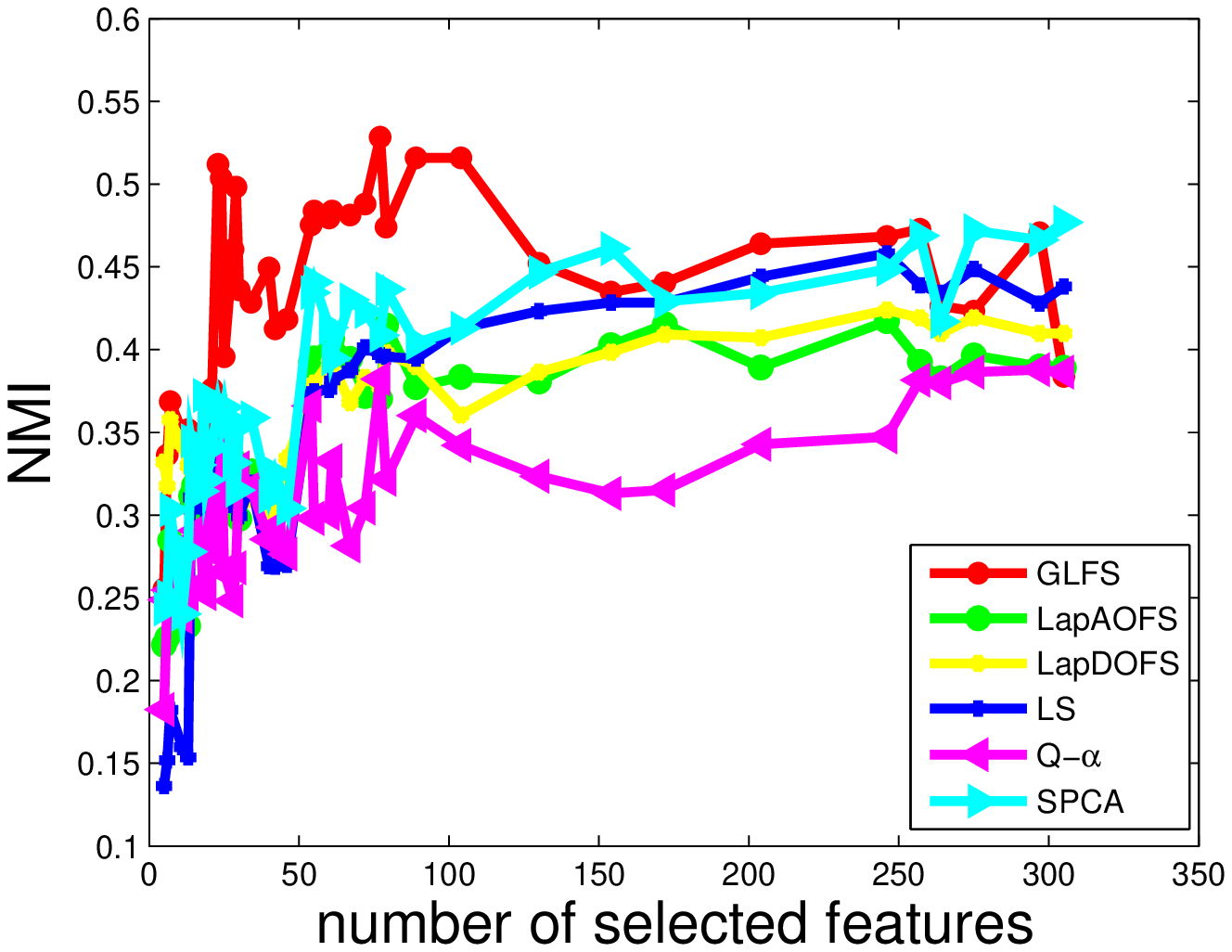}\hspace{-8mm} \\
(A)&(B)\\
\end{tabular}
\end{center}
\caption{(A) shows the number of selected features as a function of $\lambda$. (B) shows the NMI values when different numbers of features are selected.}
\label{fig:glio}
\vspace{-1mm}
\end{figure}

First, we show the pattern of sparsification of GLFS on Fig.(\ref{fig:glio})(A). We can see that, as $\lambda$ goes up, we obtain fewer features, stabilizing at about $\lambda=0.5$ corresponding to about $7$ features -- a very sparse but still a not an empty set of features, indicating that our search method converges to a nontrivial solution in practice. We perform spectral clustering~\cite{spectralclustering} and calculate the NMI values between the clustering results and the ground truth (subtypes reported in \cite{Verhaak:2010}). The comparisons are in Fig.(\ref{fig:glio})(B). In spectral clustering, we run 10 independent runs and report the average NMI value.
%

From Fig.(\ref{fig:glio})(B), GLFS appears to be a clear winner for a smaller number of features. Interestingly, our experiments show that increasing the number of features doesn't always improve the clustering results, meaning that the sparse set of fewer than a 100 features captures most of the information relevant to clustering/classification and extra features serve as added noise. Unlike GLFS other methods improve when the number of features increases.  Another interesting observation is that, LS, LapAOFS and LapDOFS are very sensitive to the patient similarity graph. Different similarity matrices can lead to very different clustering performance. Here we only report their optimal performances across many trials. Also, not shown, different initializations result in very different performances of $Q-\alpha$. This unstable behaviour was also noted in \cite{xiaofei12}. Finally, unlike in the imaging datasets, SPCA on average outperforms many competing algorithms, still falling short of the quality of the GLFS performance. Specifically, a local performance maximum at around 50 features is far below the performance of GLFS at about 25 features, meaning that GLFS captures the essence of the glioblastoma subtypes with fewer features.

\subsection{Leukaemia Classification and Gene Selection}
Since GLFS originates from the Laplacian regularized regression, it provides a natural means to perform classification. From (\ref{eqn:expW}), we can obtain the label for a test data $\mathbf{x}_{new}$:
\begin{eqnarray}
\label{eq:classifier}
y_{new} &=& \mathbf{w}^{T}\mathbf{x}_{new}\nonumber\\
&=&\mathbf{y}^T(\mathbf{X}^{\mathcal{G}})^T(\mathbf{X}^{\mathcal{G}}(\mathbf{X}^{\mathcal{G}})^T\nonumber\\
 &+& \lambda_1\mathbf{X}^{\mathcal{G}}L(\mathbf{X}^{\mathcal{G}})^T + \lambda_2I)^{-1}\mathbf{x}_{new}.
\end{eqnarray}

In this section, we test the performance of the classifier based on the chosen set of features for a well-known Leukemia microarray dataset~\cite{Golub99}. It consists of 72 samples, each with expression profiles for about 7129 genes. The task is to distinguish between the two variants of leukemia: ALL and AML. A typical training/test splitting is used(the first 38 examples as training set and the remaining 34 as the test set). To apply the classifier defined in Eqn \ref{eq:classifier}, we first coded the type of leukaemia as a 0-1 response $\mathbf{y}$. We do not perform any preprocessing (e.g. screening to reduce the number of features as in ~\cite{elasticnet}) and we use ten-fold CV to determine the parameter $\lambda$. We give our comparison results in Table \ref{tab:elasticcomparison}. For convenience, we reproduce the results from \cite{elasticnet} here in addition to our results (Golub - original classification result from \cite{Golub99},SVM - Support Vector Machines, LR - Logistic Regression, NSC - Nearest Shrunken Centroids). GLFS achieves better training accuracy and the same test accuracy as the Elastic net with fewer genes, again indicating that GLFS is robust to noise in zeroing in on the informative features.
\begin{table}[!ht]
\centering
\caption{Summary of the leukaemia classification results}
\begin{tabular}{|cccc|}
\hline
Method&Tenfold& Test& Number \\
& CV Error&Error&of Genes\\
\hline
Golub&3/38&4/34&50\\
\footnotesize{SVM}&2/38&1/34&31\\
\footnotesize{Penalized LR}&2/38&1/34&26\\
\footnotesize{NSC}&2/38&2/34&21\\
Elastic net&3/38&0/34&45\\
\bf{GLFS}&0/38&0/34&16\\
\hline
\end{tabular}
\label{tab:elasticcomparison}
\vspace{-5mm}
\end{table}

\section{Conclusion}
We have presented GLFS -- a novel unsupervised feature selection method for high dimensional noisy data. Similarly to \cite{xiaofei12} based on Laplacian regularization regression model, GLFS aims to find the sparse subset of features that are relevant to the data manifolds. Unlike \cite{xiaofei12}, GLFS optimizes exact covariance, relaxing the resulting objective function to make the method scale to very large numbers of features. We show that in practice our relaxation to $\ell_1$ coupled with early stopping of the numerical optimization significantly outperforms greedily solving $\ell_0$ of \cite{xiaofei12}. In practice GLFS is more accurate, robust and scalable than 5 other popular unsupervised methods. GLFS results in sparser solutions while significantly improving classification and clustering performance on a variety of real world datasets in the areas of object recognition and computational biology, even beating the popular Elastic Net in generalization on a supervised task. Future work includes more extensive theoretical analysis of the proposed method and refining the $\ell_1$ relaxation to make it more robust to noise.

\footnotesize
\bibliographystyle{icml2013}
\bibliography{refer}

\begin{thebibliography}{27}
\providecommand{\natexlab}[1]{#1}
\providecommand{\url}[1]{\texttt{#1}}
\expandafter\ifx\csname urlstyle\endcsname\relax
  \providecommand{\doi}[1]{doi: #1}\else
  \providecommand{\doi}{doi: \begingroup \urlstyle{rm}\Url}\fi

\bibitem[Andrew \& Gao(2007)Andrew and Gao]{Andrew:2007}
Andrew, Galen and Gao, Jianfeng.
\newblock Scalable training of l1-regularized log-linear models.
\newblock In \emph{Proceedings of the 24th international conference on Machine
  learning}, pp.\  33--40, 2007.

\bibitem[Bach \& Jordan(2004)Bach and Jordan]{spectralclustering}
Bach, Francis~R. and Jordan, Michael~I.
\newblock Learning spectral clustering.
\newblock In \emph{Advances in Neural Information Processing Systems 16}. 2004.

\bibitem[Belkin et~al.(2006)Belkin, Niyogi, and Sindhwani]{LapRLS06}
Belkin, Mikhail, Niyogi, Partha, and Sindhwani, Vikas.
\newblock Manifold regularization: A geometric framework for learning from
  labeled and unlabeled examples.
\newblock \emph{J. Mach. Learn. Res.}, 7:\penalty0 2399--2434, 2006.

\bibitem[Cai et~al.(2010)Cai, Zhang, and He]{Cai:2010}
Cai, Deng, Zhang, Chiyuan, and He, Xiaofei.
\newblock Unsupervised feature selection for multi-cluster data.
\newblock In \emph{Proceedings of the 16th ACM SIGKDD international conference
  on Knowledge discovery and data mining}, pp.\  333--342, 2010.

\bibitem[Caselles et~al.(1997)Caselles, Kimmel, and Sapiro]{activecontour}
Caselles, Vicent, Kimmel, Ron, and Sapiro, Guillermo.
\newblock {Geodesic Active Contours}.
\newblock \emph{International Journal of Computer Vision}, 22\penalty0
  (1):\penalty0 61--79, 1997.

\bibitem[Constantinopoulos et~al.(2006)Constantinopoulos, Titsias, and
  Likas]{Constantinopoulos:2006}
Constantinopoulos, Constantinos, Titsias, Michalis~K., and Likas, Aristidis.
\newblock Bayesian feature and model selection for gaussian mixture models.
\newblock \emph{IEEE Trans. Pattern Anal. Mach. Intell.}, 28\penalty0
  (6):\penalty0 1013--1018, 2006.

\bibitem[Dattorro(2011)]{Dattorro2011}
Dattorro, Jon.
\newblock \emph{Convex Optimization \& Euclidean Distance Geometry}.
\newblock Meboo Publishing USA, 2011.

\bibitem[Duda et~al.(2001)Duda, Hart, and Stork]{duda:hart:stork:2001}
Duda, R.~O., Hart, P.~E., and Stork, D.~G.
\newblock \emph{Pattern classification, second edition}.
\newblock Wiley, 2001.

\bibitem[Dy \& Brodley(2004)Dy and Brodley]{Dy:2004}
Dy, Jennifer~G. and Brodley, Carla~E.
\newblock Feature selection for unsupervised learning.
\newblock \emph{J. Mach. Learn. Res.}, 5:\penalty0 845--889, 2004.

\bibitem[Erhan et~al.(2010)Erhan, Bengio, Courville, Manzagol, Vincent, and
  Bengio]{Erhan2010}
Erhan, Dumitru, Bengio, Yoshua, Courville, Aaron, Manzagol, Pierre-Antoine,
  Vincent, Pascal, and Bengio, Samy.
\newblock Why does unsupervised pre-training help deep learning?
\newblock \emph{J. Mach. Learn. Res.}, 11:\penalty0 625--660, 2010.

\bibitem[{Fan} \& {Li}(2006){Fan} and {Li}]{Fan06}
{Fan}, J. and {Li}, R.
\newblock {Statistical Challenges with High Dimensionality: Feature Selection
  in Knowledge Discovery}.
\newblock \emph{ArXiv Mathematics e-prints}, 2006.

\bibitem[Golub et~al.(1999)Golub, Slonim, Tamayo, Huard, Gaasenbeek, Mesirov,
  Coller, Loh, Downing, Caligiuri, and Bloomfield]{Golub99}
Golub, T.~R., Slonim, D.~K., Tamayo, P., Huard, C., Gaasenbeek, M., Mesirov,
  J.~P., Coller, H., Loh, M.~L., Downing, J.~R., Caligiuri, M.~A., and
  Bloomfield, C.~D.
\newblock Molecular classification of cancer: class discovery and class
  prediction by gene expression monitoring.
\newblock \emph{Science}, 286:\penalty0 531--537, 1999.

\bibitem[Guyon et~al.(2002)Guyon, Weston, Barnhill, and Vapnik]{Guyon:2002}
Guyon, Isabelle, Weston, Jason, Barnhill, Stephen, and Vapnik, Vladimir.
\newblock Gene selection for cancer classification using support vector
  machines.
\newblock \emph{Mach. Learn.}, 46\penalty0 (1-3):\penalty0 389--422, 2002.

\bibitem[He et~al.(2005)He, Cai, and Niyogi]{LScore05}
He, Xiaofei, Cai, Deng, and Niyogi, Partha.
\newblock Laplacian score for feature selection.
\newblock In \emph{NIPS}, 2005.

\bibitem[He et~al.(2011)He, Ji, Zhang, and Bao]{xiaofei12}
He, Xiaofei, Ji, Ming, Zhang, Chiyuan, and Bao, Hujun.
\newblock A variance minimization criterion to feature selection using
  laplacian regularization.
\newblock \emph{IEEE Trans. Pattern Anal. Mach. Intell.}, 33\penalty0
  (10):\penalty0 2013--2025, 2011.

\bibitem[Liu \& Yu(2005)Liu and Yu]{liu05}
Liu, Huan and Yu, Lei.
\newblock {Toward integrating feature selection algorithms for classification
  and clustering}.
\newblock \emph{IEEE Transactions on Knowledge and Data Engineering},
  17\penalty0 (4):\penalty0 491--502, 2005.

\bibitem[Nocedal \& Wright(2000)Nocedal and Wright]{nocedal99}
Nocedal, Jorge and Wright, Stephen~J.
\newblock \emph{{Numerical Optimization}}.
\newblock Springer, 2000.

\bibitem[Robnik-\v{S}ikonja \& Kononenko(2003)Robnik-\v{S}ikonja and
  Kononenko]{Robnik2003}
Robnik-\v{S}ikonja, Marko and Kononenko, Igor.
\newblock Theoretical and empirical analysis of relieff and rrelieff.
\newblock \emph{Mach. Learn.}, 53\penalty0 (1-2):\penalty0 23--69, 2003.
\newblock ISSN 0885-6125.

\bibitem[TCGA(2008)]{TCGA}
TCGA.
\newblock Comprehensive genomic characterization defines human glioblastoma
  genes and core pathways.
\newblock \emph{Nature}, 455\penalty0 (7216):\penalty0 1061--8, Oct 2008.
\newblock \doi{10.1038/nature07385}.

\bibitem[Teodoro et~al.(2003)Teodoro, Phillips, and Kavraki]{miguel03}
Teodoro, Miguel~L., Phillips, George~N., and Kavraki, Lydia~E.
\newblock Understanding protein flexibility through dimensionality reduction.
\newblock \emph{Journal of Computational Biology}, 10:\penalty0 617--634, 2003.

\bibitem[Tibshirani(1994)]{Tibshirani94}
Tibshirani, Robert.
\newblock Regression shrinkage and selection via the lasso.
\newblock \emph{Journal of the Royal Statistical Society, Series B},
  58:\penalty0 267--288, 1994.

\bibitem[Verhaak et~al.(2010)Verhaak, Hoadley, Purdom, Wang, Qi, Wilkerson,
  Miller, Ding, Golub, Mesirov, Alexe, Lawrence, O'Kelly, Tamayo, Weir,
  Gabriel, Winckler, Gupta, Jakkula, Feiler, Hodgson, James, Sarkaria, Brennan,
  Kahn, Spellman, Wilson, Speed, Gray, Meyerson, Getz, Perou, Hayes, and
  {Cancer Genome Atlas Research Network}]{Verhaak:2010}
Verhaak, Roel G~W, Hoadley, Katherine~A, Purdom, Elizabeth, Wang, Victoria, Qi,
  Yuan, Wilkerson, Matthew~D, Miller, C~Ryan, Ding, Li, Golub, Todd, Mesirov,
  Jill~P, Alexe, Gabriele, Lawrence, Michael, O'Kelly, Michael, Tamayo, Pablo,
  Weir, Barbara~A, Gabriel, Stacey, Winckler, Wendy, Gupta, Supriya, Jakkula,
  Lakshmi, Feiler, Heidi~S, Hodgson, J~Graeme, James, C~David, Sarkaria,
  Jann~N, Brennan, Cameron, Kahn, Ari, Spellman, Paul~T, Wilson, Richard~K,
  Speed, Terence~P, Gray, Joe~W, Meyerson, Matthew, Getz, Gad, Perou,
  Charles~M, Hayes, D~Neil, and {Cancer Genome Atlas Research Network}.
\newblock Integrated genomic analysis identifies clinically relevant subtypes
  of glioblastoma characterized by abnormalities in pdgfra, idh1, egfr, and
  nf1.
\newblock \emph{Cancer Cell}, 17\penalty0 (1):\penalty0 98--110, Jan 2010.
\newblock \doi{10.1016/j.ccr.2009.12.020}.

\bibitem[Wolf \& Shahua(2003)Wolf and Shahua]{wolf03}
Wolf, L. and Shahua, A.
\newblock {Feature Selection for Unsupervised and Supervised Inference: the
  Emergence of Sparsity In Weighted Based}.
\newblock 2003.

\bibitem[Xu et~al.(2003)Xu, Liu, and Gong]{Xu03}
Xu, Wei, Liu, Xin, and Gong, Yihong.
\newblock Document clustering based on non-negative matrix factorization.
\newblock In \emph{Proceedings of the 26th annual international ACM SIGIR
  conference on Research and development in informaion retrieval}, pp.\
  267--273, 2003.

\bibitem[Yang et~al.(2007)Yang, Wright, Ma, and Sastry]{Yang:EECS-2007-99}
Yang, Allan~Y., Wright, John, Ma, Yi, and Sastry, S.~Shankar.
\newblock Feature selection in face recognition: A sparse representation
  perspective.
\newblock Technical Report UCB/EECS-2007-99, EECS Department, University of
  California, Berkeley, Aug 2007.
\newblock URL
  \url{http://www.eecs.berkeley.edu/Pubs/TechRpts/2007/EECS-2007-99.html}.

\bibitem[Zou \& Hastie(2005)Zou and Hastie]{elasticnet}
Zou, Hui and Hastie, Trevor.
\newblock Regularization and variable selection via the elastic net.
\newblock \emph{Journal of the Royal Statistical Society Series B}, 67\penalty0
  (2):\penalty0 301--320, 2005.

\bibitem[Zou et~al.(2004)Zou, Hastie, and Tibshirani]{Zou04sparseprincipal}
Zou, Hui, Hastie, Trevor, and Tibshirani, Robert.
\newblock Sparse principal component analysis.
\newblock \emph{Journal of Computational and Graphical Statistics},
  15:\penalty0 2006, 2004.

\end{thebibliography}
\end{document}